\PassOptionsToPackage{doi=false,isbn=false,url=false}{bibtex}
\documentclass[letterpaper, 10 pt, conference]{ieeeconf} 

\IEEEoverridecommandlockouts                              

\usepackage{graphicx} 
\graphicspath{ {./figures/} }
\usepackage{amsmath} 
\usepackage{amssymb}
\usepackage{algorithm}
\usepackage[noend]{algpseudocode}
\usepackage{xcolor}
\definecolor{tumblue}{RGB}{0,101,189}

\usepackage{stfloats}
\usepackage{setspace}
\usepackage{tabularx}
\usepackage{booktabs}
\usepackage[T1]{fontenc}
\usepackage[scaled=0.85]{beramono}
\usepackage{listings}
\usepackage{caption}
\usepackage{cleveref}
\usepackage{url}

\Crefname{lstlisting}{Listing}{Listings}
\usepackage{listings}
\usepackage{multirow}
\lstdefinestyle{sparqlstyle}{
  language=OCL,
  basicstyle=\footnotesize,
  stepnumber=1,
  numbersep=10pt,
  tabsize=2,
  showspaces=false,
  breaklines=true
}

\usepackage[nonumberlist,nopostdot,acronym,toc]{glossaries}
\newacronym{bt}{BT}{Behavior Tree}
\newacronym{fsm}{FSM}{Finite State Machine}
\newacronym{llm}{LLM}{Large Language Model}
\newacronym{6dof}{6DOF}{Six Degrees of Freedom}
\newacronym{hrc}{HRC}{Human-Robot Collaboration}
\newacronym{asr}{ASR}{Automatic Speech Recognition}
\newacronym{hra}{HRA}{Human Resource Agent}

\usepackage{balance} 

\makeatletter
\newcommand\fs@betterruled{%
  \def\@fs@cfont{\bfseries}\let\@fs@capt\floatc@ruled
  \def\@fs@pre{\vspace*{5pt}\hrule height.8pt depth0pt \kern2pt}%
  \def\@fs@post{\kern2pt\hrule\relax}%
  \def\@fs@mid{\kern2pt\hrule\kern2pt}%
  \let\@fs@iftopcapt\iftrue}
\floatstyle{betterruled}
\restylefloat{algorithm}
\makeatother

\author{Jonghan Lim$^{1}$, Sujani Patel$^{2}$, Alex Evans$^{3}$, John Pimley$^{1}$, Yifei Li$^{2}$, and Ilya Kovalenko$^{1,2}$
\thanks{The authors [1] are with the Department of Industrial and Manufacturing, [2] are with the Department of Mechanical Engineering, and [3] are with the Department of Computer Science and Engineering, at the Pennsylvania State University, State College, USA
        (e-mail: \{jxl567; spp22; ade5221; jap6581; ybl5717;  iqk5135\}@psu.edu).
        }%
}

\title{\LARGE \bf
Enhancing Human-Robot Collaborative Assembly in Manufacturing Systems Using Large Language Models
}

\begin{document}

\setlength{\textfloatsep}{5pt}
\maketitle
\thispagestyle{empty}
\pagestyle{empty}

\begin{abstract}
The development of human-robot collaboration has the ability to improve manufacturing system performance by leveraging the unique strengths of both humans and robots. On the shop floor, human operators contribute with their adaptability and flexibility in dynamic situations, while robots provide precision and the ability to perform repetitive tasks. However, the communication gap between human operators and robots limits the collaboration and coordination of human-robot teams in manufacturing systems. Our research presents a human-robot collaborative assembly framework that utilizes a large language model for enhancing communication in manufacturing environments. The framework facilitates human-robot communication by integrating voice commands through natural language for task management. A case study for an assembly task demonstrates the framework's ability to process natural language inputs and address real-time assembly challenges, emphasizing adaptability to language variation and efficiency in error resolution. The results suggest that large language models have the potential to improve human-robot interaction for collaborative manufacturing assembly applications.
\end{abstract}

\section{Introduction}
\label{sec:introduction}

Advances in robotics technology have significantly enhanced manufacturing efficiency, resulting in cost reductions and increased productivity~\cite{paryanto2015reducing, wang2020overview}. While robots can execute rapid, accurate, and repetitive tasks that demand heavy-duty efforts in manufacturing settings, they lack the capability for adaptation and versatility of human operators~\cite{wang2020overview}. Therefore, the importance of \gls{hrc} is growing as humans and robots complement each other skills and capabilities.
\gls{hrc} refers to the interaction and cooperation between human operators and robotic systems within shared workspace~\cite{franklin2020collaborative}. Prior works have used \gls{hrc} frameworks to improve ergonomics in manufacturing settings and safe human-robot interaction without compromising productivity~\cite{zanchettin2015safety, pearce2018optimizing}. Transitioning from tasks involving large components, such as loading, placing, and unloading, to the complex assembly of smaller components such as printed circuit boards, the collaboration between human operators and robots significantly enhances both the efficiency and safety of production lines~\cite{javaid2021substantial, bogner2018optimised}. 

Further advancing \gls{hrc} in manufacturing systems by improving the interaction of human operators and robots presents significant challenges. Specifically, interaction with robots induces psychological stress and tension among operators due to language barrier~\cite{korner2019perceived}. In contemporary manufacturing systems, human operators require extensive pre-service training and complex code development to ensure accurate and safe production with the robots~\cite{matheson2019human}. These difficulties highlight the need to develop human-robot communication systems that enable interaction between humans and robots without extensive robotics training \textit{(C1)}. 
Another challenge involves the need for greater flexibility and adaptability in human-robot interaction. As the number of interactions between humans and robots on the shop floor increases, there is a higher chance of encountering unexpected changes and errors. Thus, \gls{hrc} needs to be more adaptable to changes and errors during the manufacturing assembly process \textit{(C2)}. Additionally, human-robot collaborative assembly applications must integrate advanced technologies with a human-centric design to improve communication and usability \textit{(C3)}.

\glspl{llm} have recently been introduced in the AI community to enhance natural language understanding and generation capabilities. These can be extended to improve human-robot interaction in the manufacturing facility. Models such as OpenAI's GPT-3~\cite{floridi2020gpt} and GPT-4~\cite{achiam2023gpt}, have shown proficiency in processing, understanding, and communicating in natural language. The integration of \glspl{llm} facilitates natural language communication between humans and robots. Using voice interaction for this communication enhances collaboration and operator safety in dynamic workspaces.
The main contributions of this work are:
(1) the use of \glspl{llm} for interpreting natural language to allow operators to coordinate with robotic arms,
(2) a framework for system integration of voice commands, robotic arms, and vision systems to facilitate \gls{hrc} in an assembly task, enhancing operational flexibility, and
(3) an adaptation to task errors and obstacles through human-robot communication in a manufacturing environment.

The rest of the manuscript is organized as follows. 
\Cref{sec:background} reviews related work and identifies gaps in \gls{hrc} for manufacturing assembly. 
\Cref{sec:framework} describes the proposed framework that integrates \glspl{llm} with voice commands, robotic arm, and sensors. 
\Cref{sec:casestudy} showcases a case study and evaluates the framework's performance. 
Finally, \Cref{sec:conclusion} provides conclusions and future work.

\section{Background}
\label{sec:background}

\subsection{Human-Robot Collaboration in Manufacturing}

Recently, a wide range of methods has been developed in the field of \gls{hrc} to improve the safety and efficiency of interactions between humans and robots. For instance, Fernandez et al.~\cite{de2017multimodal} developed a dual-arm robotic system with multisensor capabilities for safe and efficient collaboration, integrating gesture recognition. Additionally, Wei et al.~\cite{wei2021vision} gather information on human operators and the environment using RGB-D videos and predict human intentions based on deep-learning methods. 

Some approaches have utilized multi-modal strategies, including natural language, to improve communication for \gls{hrc} in manufacturing. For example, Liu et al.~\cite{liu2018towards} focus on integrating various modalities, including speech commands, hand motions, and body motions to improve \gls{hrc}. While it addresses speech command recognition using deep learning models, the interaction is limited to basic speech command recognition and does not focus on context-aware communication.
Additionally, Wang et al.~\cite{wang2018human, wang2018facilitating} employ a teaching-learning model that includes natural language instructions to predict human intentions and facilitate collaboration. While this model uses natural language for multimodal processing, it does not emphasize utilizing language variations in interaction.

Previous research have introduced methods using environmental data to enhance safety and efficiency and natural language to improve \gls{hrc} in manufacturing. However, there is limited research on human-robot collaborative assembly that effectively integrates natural language capabilities for context-aware communication and handling language variations. We aim to integrate an \gls{llm}-based approach to improve communication between humans and robots. Our approach is an initial step towards enhancing human-robot interaction by combining existing technologies, such as computer vision and \gls{llm}, to leverage human flexibility and robot precision in manufacturing.

\subsection{Large Language Models for Robots}
Recent research has explored integrating \glspl{llm} in robotic applications to enhance human-robot interaction and robot functionalities~\cite{huang2022language, singh2022progprompt, Lin_2023, mees2023grounding, zhang2024building, ren2023robots}. This includes converting high-level human instructions into tasks robots can execute. Huang et al.~\cite{huang2022language} used \glspl{llm}, such as GPT-3 and CODEX, to convert high-level human instructions into executable actions for robots. Similarly, Singh et al.~\cite{singh2022progprompt} developed a method called ProgPrompt, which generates Python code for task planning that accounts for environmental conditions and pre-programmed actions. Furthermore, Lin et al.~\cite{Lin_2023} developed a framework for taking natural language instructions and then developing task and motion plans.
Other works in robotics have utilized \glspl{llm} to maximize the robot's capability to comprehend and interact with its environment to resolve errors. Mees et al~\cite{mees2023grounding} combined natural language input with visual information to enhance task completion ability. Zhang et al.~\cite{zhang2024building} utilizes \gls{llm} to address multi-agent cooperation challenges, resolving issues related to decentralized control, raw sensory data processing, and efficient task execution in embodied environments. KnowNo~\cite{ren2023robots} improved robot autonomy and efficiency by guiding robots in seeking human help when faced with uncertainty. 

These studies show how integrating \glspl{llm} into robotic applications improves human-robot interaction and robot capabilities. 
However, existing studies have not addressed common challenges found in the manufacturing environment, such as matching natural language to structured task sequences and managing the timing of commands (e.g., identifying when to request operator assistance in the process).
We aim to bridge the gap by improving the integration of natural language for human-robot interaction for an assembly process through an \gls{llm}-based framework.

\section{framework}
\label{sec:framework}

\begin{figure*}[t]
\smallskip
\smallskip
    \centering
    \captionsetup{belowskip=-17pt} 
    \includegraphics[width=.98\textwidth]{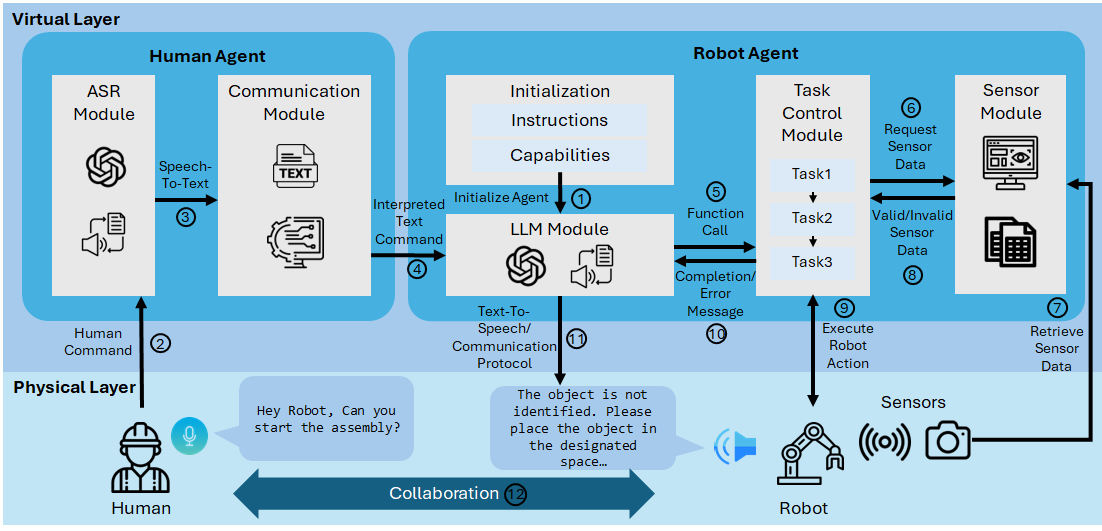}
    \caption{Human-Robot Collaborative Assembly Framework Using \gls{llm}}
    \label{fig:framework}
\end{figure*}

The proposed framework, presented in~\Cref{fig:framework}, is developed for a human-robot collaborative assembly within the manufacturing environment. The framework is designed to facilitate interaction between a human operator and a robot for an assembly process.

In this framework, a human can initialize and provide a set of tasks \(T = \{t_1, t_2, \ldots, t_n\}\) as the objectives to be achieved by the robot. 
While $T$ is predefined, the system can be extended to integrate new tasks based on human input.
The robot's capabilities \(C = \{c_1, c_2, \ldots, c_l\}\) represent the specific actions programmed to perform within an assembly process. If the task matches the robot's capabilities, the robot should execute the specified task $t_i$. The task, $t_i$, is broken down into subtasks (\({t_{i1}, t_{i2}, \ldots, t_{ik}}\)) that outline the sequential steps for completion. The framework has to track the last completed subtask, denoted as $t_{i\text{c}}$, where $1 \leq c \leq k$ indicates the completed process within ${t_{i1}, t_{i2}, \ldots, t_{ik}}$. When an error is resolved by a human operator, the robot resumes from $t_{i\text{c}+1}$, ensuring continuous task progression. To enable this human-robot collaborative assembly process, this framework is structured into two primary layers: the physical layer and the virtual layer, bridging the gap between human inputs and robot actions.

\subsection{Physical Layer}
\label{subsec:physical}

The physical layer facilitates human-robot interaction within a shared space, based on data from the virtual layer. The physical consists of three primary components: human commands, robot actions, and sensor data. Human commands involve the operator providing voice instructions to control robot actions. Robot actions are based on a predefined set of $T$. Sensor data is utilized to monitor environmental conditions. This data ensures the action adapts to changes in the workspace (e.g. position, orientation, or availability of parts).

When a predefined event or error $e_i \in E$ is detected, the robot alerts the human operator using a communication protocol. The \gls{llm} module converts $e_i$ into a natural language message \(M_{\text{e}_i}(t_i)\), delivered via text-to-speech technology. Different instances of \(e_i\) can lead to varied message \(M_{\text{e}_i}(t_i)\) depending on the task \(t_i\) error. After receiving and understanding \(M_{\text{e}_i}(t_i)\), the human operator resolves the corresponding $e_i$ and commands the robot to resume $t_i$.

\subsection{Virtual Layer}
\label{subsec:virtual}

The virtual layer serves to facilitate communication. This layer stores the system's functionality to enable interaction between human commands and robot actions. The virtual layer of the system consists of two main agents: a human agent and a robot agent.

\subsubsection{Human Agent}
\label{subsubsec:human}
We have integrated a human agent, often referred to as a human command interpreter, to enhance the human-robot interaction based on the approach by Zhang et al.~\cite{zheng2018integrating}. The goal of the human agent is to process the human commands in voice instructions into a text that the robot agent can understand. This agent ensures that the human intentions are transferred to the robot without manual programming. The human agent contains an \gls{asr} module responsible for processing the human command voice data, which converts this data into text that the robot agent can process. The communication module in the human agent serves as a bridge, transmitting commands and information to the robot.

\subsubsection{Robot Agent}
\label{subsubsec:robot}

The robot agent interprets and executes the set of manufacturing tasks $T$ based on the voice commands received from the human operator. This process is facilitated by various functional modules:

\begin{itemize}
    \item \textbf{Initialization Module:} The initialization of the robot agent, provides basic operational guidelines and task execution protocols. This process involves programming the robot with instructions to perform tasks $T$ within its capabilities $C$ and to communicate errors or request assistance from human operators when necessary. The following initial prompt is given to the robot agent:
    
    \begin{quote}
    {\footnotesize
    \texttt{"You are a robot agent in a human-robot collaborative assembly system designed to assist in tasks and respond to commands. Upon receiving a request within your capability range, execute the service. In the event of encountering errors, request assistance from a human operator for error correction, providing clear and understandable explanations."}
    }
    \end{quote}
    
    This initialization prompt defines the robot agent's role in responding to commands. It also establishes a protocol for seeking assistance and communicating errors to human operators for task completion. Following the initial setup of the robot agent, specific capabilities $C$ are defined, such as specific assembly tasks. 
    
    \item \textbf{\gls{llm} Module:} An \gls{llm} interprets human commands into tasks $t_i \in T$ using initialization prompts and detailed functional information. By processing commands based on context, such as the last completed subtask and any encountered errors, the \gls{llm} can automatically detect and suggest the next task. This process allows the system to match the command with the predefined functional capabilities $C$, ensuring the intended task is executed. The \gls{llm} module also communicates any detected issues from the task control module into understandable natural language for the human operators, facilitating \gls{hrc} for error resolution. 
    Using an \gls{llm} to process static error messages can generate contextual information, helping human operators understand and resolve issues.
            
    \item \textbf{Sensor Module:} The sensor module processes data from the physical layer such as position, orientation, and part availability. This information is utilized for adjusting robot actions required for executing manufacturing tasks $T$. For example, the module provides component positions and orientations for the task control module. This enables accurate robotic adjustments for assembly or initiates error notifications for missing components, ensuring effective task execution and error management.    
    
    \item \textbf{Task Control Module:} The task control module directs robot actions to fulfill task $t_i$ within the set $T$, based on the functional capabilities $C$. The task control module adjusts robot actions to environmental conditions obtained from the sensor module. 
    Additionally, this module plays a role in managing errors by verifying sensor data and communicating detected issues through the \gls{llm} module to facilitate resolution by the human operator.
    
\end{itemize}

\subsection{Human-Robot Collaborative Assembly Workflow}
\label{subsec:workflow}

\begin{figure}[t]
\smallskip
\smallskip
    \centering
    \captionsetup{belowskip=-5pt} 
    \includegraphics[width=.48\textwidth]{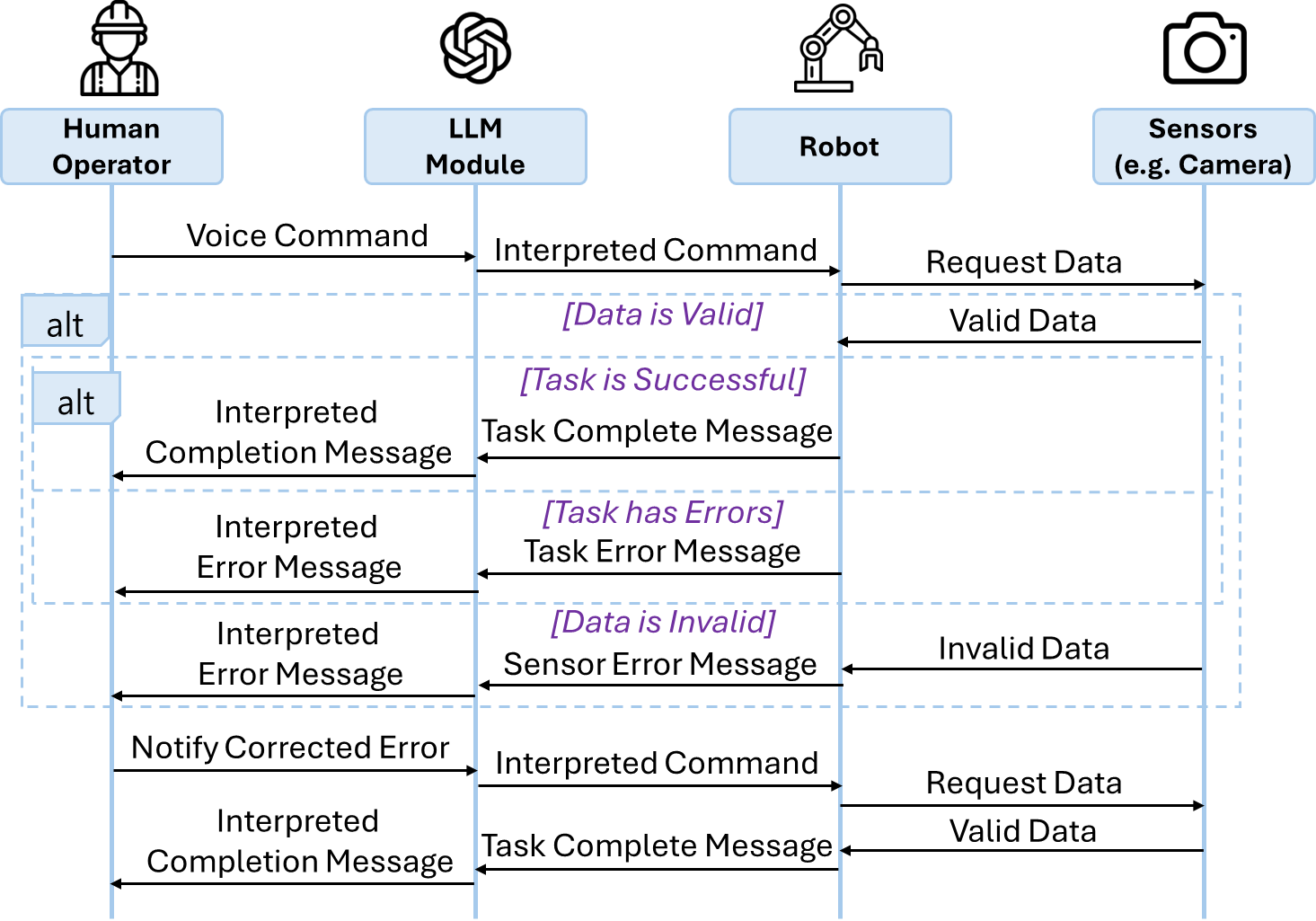}
    \caption{Sequence Diagram for Human-Robot Collaborative Assembly in Manufacturing Systems}
    \label{fig:sequence}
\end{figure}

The overall workflow is shown in a sequence diagram in ~\Cref{fig:sequence}, illustrating the human-robot collaborative assembly process. The diagram depicts how voice commands from the human operator are processed by the \gls{llm} module to guide robot actions.

The process starts with the operator giving a voice command, converted by the \gls{llm} module into a discrete set of tasks $T$ for the robot. The robot then requests sensor data to execute $t_i$. If the data is valid, the robot proceeds to execute the assigned $t_i$. The sensor module determines data validity by comparing detected parameters to predefined criteria. Successful execution results in a completion message \(M_{\text{c}}(t_i)\) to the operator via the \gls{llm} Module.

If the data is invalid, or $t_i$ has any errors, the robot generates an error message \(M_{\text{e}_i}(t_i)\) via the \gls{llm} module, aiming to inform the human operator of the specific error and its occurrence within subtask $t_{i\text{c}+1}$ for efficient resolution. Following the error identification and correction by the human operator, a new command by the human operator is issued to the robot. The robot then resumes task execution at $t_i$, starting from the interrupted subtask $t_{i\text{c}+1}$, based on the new sensor data. This procedure is repeated until $t_i$ is completed.

\section{Case Study}
\label{sec:casestudy}

\begin{figure}[t]
\smallskip
\smallskip
    \centering
    \captionsetup{belowskip=-5pt} 
    \includegraphics[width=.48\textwidth]{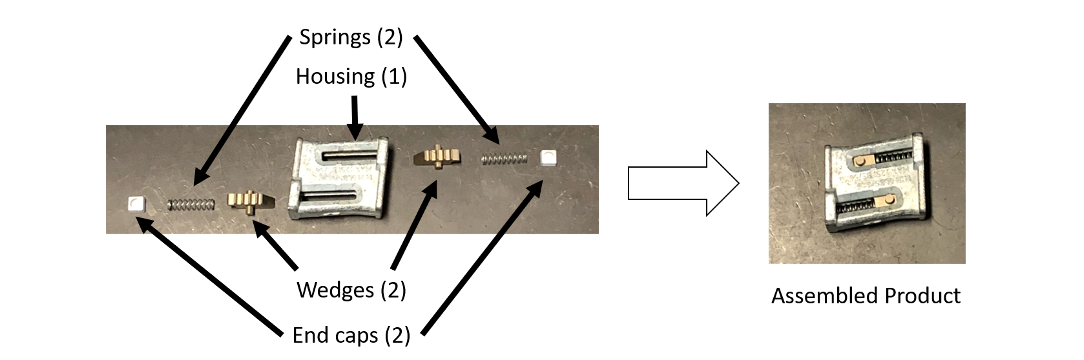}
    \caption{Cable Shark Assembly}
    \label{fig:assembly}
\end{figure}

The proposed framework was tested in a manufacturing assembly manufacturing cell. The goal of the manufacturing cell is to assemble a cable shark product. The cable shark assembly contains (1) a housing, (2) a wedge, (3) a spring, (4) and an end cap. The exploded cable shark assembly and the final assembled product are shown in~\Cref{fig:assembly}. This assembly process task $t_1$ include four sequential subtasks: housing assembly ${t_{11}}$, wedge assembly ${t_{12}}$, spring assembly ${t_{13}}$, and end cap assembly ${t_{14}}$. 

The physical setup uses a \gls{6dof} UFactory xArm with a UFactory 2-finger gripper as the end effector~\cite{UFACTORY}. The base of the robotic arm remains fixed at the center, accompanied by a mat on one side, while the parts are assembled on the adjacent side. Two Intel RealSense Depth Camera D435 cameras were used as visual sensors in the surrounding area, one focused on the mat area for pre-assembly component placement, and the other camera on the assembling process.  The conversion between camera and robot arm coordinates is accomplished through the coordinate transformation. We used a fixed z-axis value for the robotic arm to pick up camera-identified parts. In future work, we aim to make this z-axis value adaptable.

\subsection{\gls{llm} and \gls{asr} Module}
\label{subsec:llm-module}

This section outlines how \gls{llm} and the \gls{asr} module, were implemented within the system. The communication aspect within the system is enabled by OpenAI's speech-to-text and text-to-speech models. The OpenAI's transcription model `whisper1'~\cite{speechtotext} is implemented to transform speech-to-text, ensuring that human voice instructions are accurately captured. To enable the communication protocol to provide a verbal response from the robot, OpenAI's text-to-speech model `tts-1'~\cite{texttospeech} is utilized to generate audio responses. This approach enhances and allows for a clear exchange of information between the human and the robot.

The \gls{llm} module leverages OpenAI's pre-trained GPT-4.0~\cite{achiam2023gpt}.
The \gls{llm} module converts instructions from the human operator into tasks $T$ executable by the robot, utilizing its capabilities $C$ through OpenAI's \textit{function calling} feature~\cite{function2024calling}. This module ensures that the robot actions match the human operator commands, allowing for adaptive and responsive task management within the manufacturing environment. 

\subsection{Sensor Module: Vision System}
\label{subsec:vision}

\begin{figure}[t]
    \centering
    \captionsetup{belowskip=-5pt} 
    \includegraphics[width=.28\textwidth]{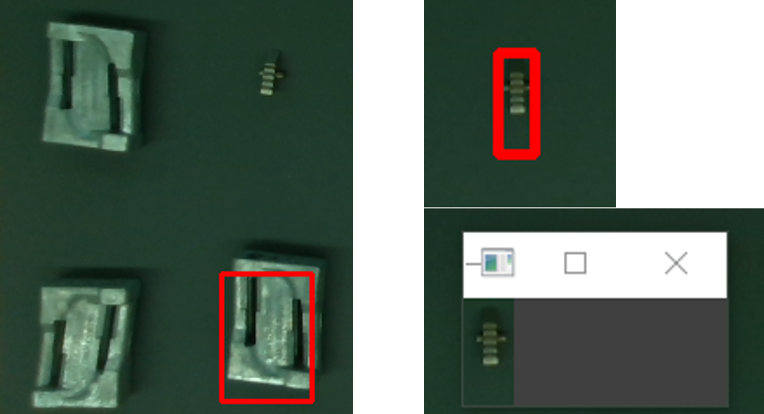}
    \caption{Feature Extraction Method with the Vision System}
    \label{fig:identification}
\end{figure}

We incorporated a vision system as our sensor module. During the collaborative assembly process, the vision system provides feedback on environmental data to the task control module. This enables precise object detection and identification of object orientation, facilitating task execution and error management. To enable object detection, Yolov5, a computer vision model, is utilized~\cite{Jocher_YOLOv5_by_Ultralytics_2020}. Custom Yolov5 models are trained using a dataset of images of individual parts (i.e. housing, wedge, spring, end cap), with manual bounding box annotations around each part in the images. The vision system these annotations to identify the parts, as shown in~\Cref{fig:identification}. If a part is detected, the coordinates of the top-left and bottom-right corners of the bounding box are retrievable. Once these coordinates are obtained, the x and y coordinates of these points are utilized to determine the object's midpoint. This midpoint serves as the target point for the robotic arm to pick up the part. Following the successful identification, the valid sensor data is transferred to the task control module for guiding robot action. For inaccurately detected parts, the sensor module transfers invalid data to the task control module to notify the human operator via \gls{llm}.

\subsection{Task Control Module: Assembly Task}
\label{subsec:taskcontrol}

\begin{figure}[t]
\smallskip
\smallskip
    \centering
    \captionsetup{belowskip=-5pt} 
    \includegraphics[scale=0.7]{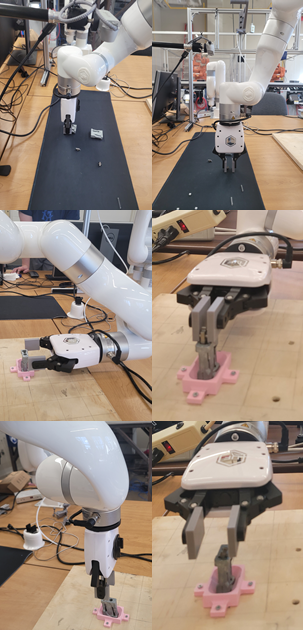}
    \caption{Cable Shark Assembly Process}
    \label{fig:process}
\end{figure}

\begin{table*}[h]
\smallskip
\smallskip
\centering
\caption{Language Variations for Task Instructions}
\label{table:language_variations}
\resizebox{\textwidth}{!}{
\begin{tabular}{|l|l|l|}
\hline
\textbf{Scenario 1} & \textbf{Instruction Type} & \textbf{Instruction} \\
\hline
\multirow{3}{*}{Scenario 1: Component Overlap} & Specific & ["Overlap resolved. Proceed with the task."], ["Problem is..."]\\ \cline{2-3}
 & Moderately Specific & ["I've placed the components correctly."], ["I've sorted out the..."]\\ \cline{2-3}
 & Least Specific & ["Fixed."], ["Done."], ["Completed."], ["Handled."], ["Adjusted."] \\
\hline
\multirow{3}{*}{Scenario 2: Incorrectly Assembled Part} & Specific & ["Correction is made. Resume the task."], ["The wedge is..."] \\ \cline{2-3}
 & Moderately Specific & ["I've fixed the issue with the wedge."], ["I've placed the wedge..."] \\ \cline{2-3}
 & Least Specific & ["Fixed."], ["Done."], ["Addressed."], ["All set."], ["Under control."] \\
\hline
\multirow{3}{*}{Scenario 3: Missing Component} & Specific & ["I've placed the spring component. Please proceed."].["Spring..."] \\ \cline{2-3}
 & Moderately Specific & ["I've fixed the issue with the spring."], ["The spring component..."] \\ \cline{2-3}
 & Least Specific & ["Fixed."], ["Done."], ["Managed."], ["Handled"], ["Settled."] \\
\hline
\end{tabular}
}
\vspace{-\baselineskip}
\end{table*}

\begin{table}[htbp]
\centering
\caption{Success Rates for Language Variations}
\label{table:success_rate}
\setlength{\extrarowheight}{2pt}
\small
\begin{tabularx}{\columnwidth}{|p{0.3\columnwidth}|X|l|}
\hline
\textbf{Scenario} & \textbf{Instruction Category} & \textbf{Success Rate} \\
\hline
\multirow{3}{0.3\columnwidth}{Scenario 1: Component Overlap} & Specific & 100\% \\
\cline{2-3}
 & Moderately Specific & 73\% \\
\cline{2-3}
 & Least Specific & 27\% \\
\hline
\multirow{3}{0.3\columnwidth}{Scenario 2: Incorrectly Assembled Part} & Specific & 93\% \\
\cline{2-3}
 & Moderately Specific & 87\% \\
\cline{2-3}
 & Least Specific & 53\% \\
\hline
\multirow{3}{0.3\columnwidth}{Scenario 3: Missing Component} & Specific & 100\% \\
\cline{2-3}
 & Moderately Specific & 67\% \\
\cline{2-3}
 & Least Specific & 27\% \\
\hline
\end{tabularx}
\vspace{-\baselineskip}
\end{table}

The Task Control Module executes tasks from human commands interpreted by the \gls{llm} and manages errors in a human-robot collaborative assembly system. The task control module verifies sensor data from the vision system. For instance, if the data is valid, the task control module proceeds; if not, the task control module sends error details (e.g. missing component) to the \gls{llm} module to convert the information into natural language. This process activates the communication protocol, allowing the human operator to recognize and address the error. 
To enable the assembly process, the task control module integrates various functionalities, including coordinate transformation and position and rotation control, based on the sensor data obtained from the vision system. These functions enable the execution of tasks interpreted by the \gls{llm}. The orientation of detected objects is determined from the bounding box dimensions, assuming the object's angle is perpendicular or parallel to the x-axis of the camera vision. The cable shark assembly process is shown in~\Cref{fig:process}.


\subsection{Case Study Results}
\label{subsec:casestudy-setup}

The proposed framework was integrated into the assembly system to study the effect of integrating LLMs with a knowledgeable operator\footnote{See supplementary video file for a demonstration of the human-robot collaborative assembly.}.
The operator knew the assembly process and was able to converse.
Three scenarios were used to evaluate the proposed framework:

\noindent
\textbf{Scenario 1}: The system detects an overlap of the housing components and requests human intervention.

\noindent
\textbf{Scenario 2}: If a wedge component is incorrectly assembled, the robot halts and requests human correction.

\noindent
\textbf{Scenario 3}: A missing spring component is detected and the robot requests the human operator to place the spring component.

Based on these scenarios, the evaluation assessed the system's proficiency in interpreting and performing tasks from human commands, considering the linguistic and human diversity. Variations in phrasing and terminology across different commands were cataloged, as presented in \Cref{table:language_variations}, to validate the system's adaptability to diverse linguistic inputs. Instructions were classified into three distinct categories: specific, moderately specific, and least specific. Five unique variations of each instruction type were provided across three distinct tasks. To ensure reliability, each language variation scenario presented in~\Cref{table:language_variations} is conducted with three repetitions.

\begin{figure}[t]
    \centering
    \captionsetup{belowskip=-5pt} 
    \includegraphics[width=.48\textwidth]{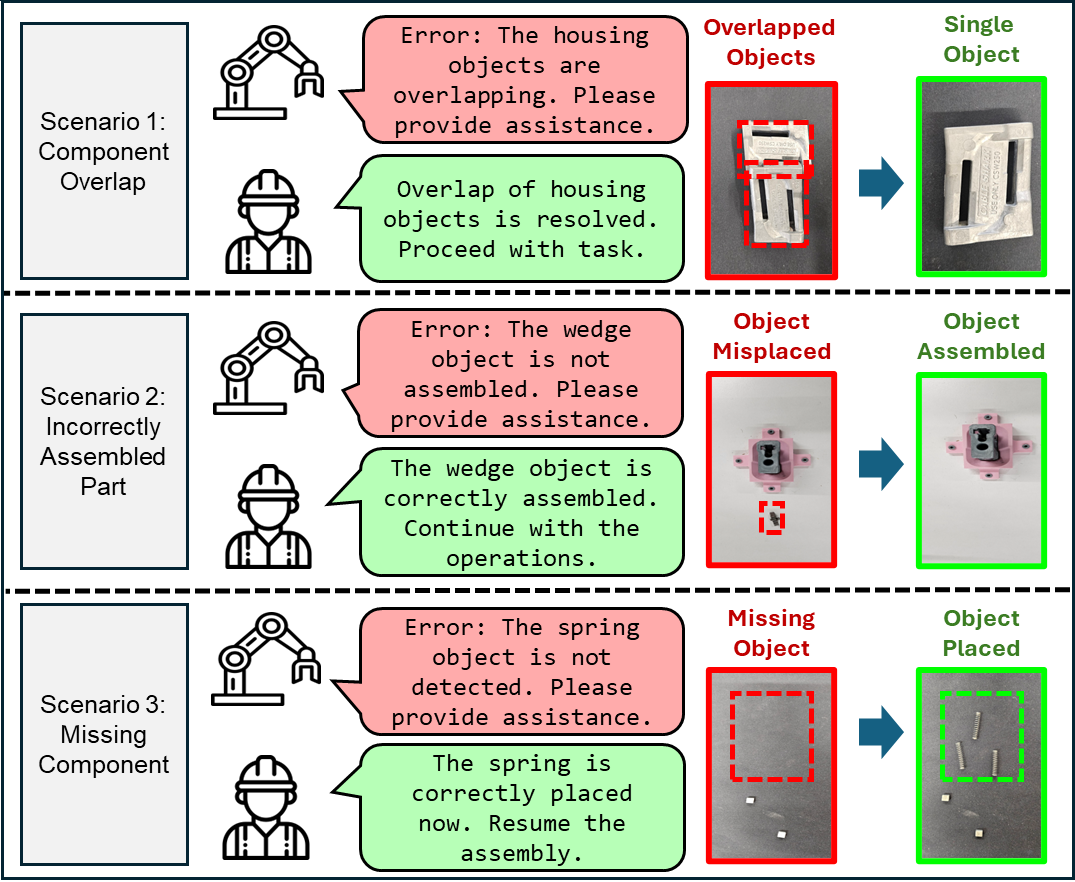}
    \caption{Case Study Communication Results for Each Scenario}
    \label{fig:casestudy-scenarios}
\end{figure}

The results highlight the effectiveness of integrating \gls{llm} into the human-robot collaborative assembly framework. \Cref{fig:casestudy-scenarios} showcases the human-robot communication using the vision system, outlined in~\Cref{subsec:vision}, to ensure task completion. In scenario 1, the robot detects overlapping components which signal to human operators for help. Upon resolution, the human operator prompts task continuation. Similarly in scenario 2, the robot identifies a misassembled wedge, notifying the human operator, and then the human commands the robot to continue after manually assembling the wedge. For Scenario 3, the robot flags a missing spring component, requesting human intervention, and the human instructs the robot to resume the assembly once the issue is fixed.

The evaluation assessed the system's capability to understand and execute commands with varied language expressions. Success rate refers to the percentage of the robot correctly understanding and executing the given instruction without requiring additional clarification. For tasks with specific instructions, success rates were high, averaging approximately 98\%. The rates decreased to 76\% with moderately specific instructions and dropped further to 36\% with least specific instructions. The success rates are detailed in~\Cref{table:success_rate}. This data suggests a positive correlation between instruction type and task execution success.

\subsection{Case Study Discussion and Limitations}

The case study evaluated the system’s ability to facilitate collaboration between humans and robots and showcases how the integration of \glspl{llm} can lead to efficient and flexible manufacturing processes. 
The results show that as instructions become less specific, robot performance significantly decreases, indicating the need for well-defined commands. For example, the ambiguous command "Correction is made. Resume the operations." failed due to lack of context and explicit task reference.
These results highlight the limitations of the proposed framework and suggest an area for improvement. One such improvement involves introducing feedback mechanisms that allow the system to ask for clarification when instructions are unclear, refining the interaction between humans and robots. Another approach is to fine-tune the \gls{llm}~\cite{xia2024leveraging} or integrate a knowledge base~\cite{lim2023ontology} with specific manufacturing processes to improve its ability to handle instructions and provide accurate responses. Commands containing keywords like 'fixed' or 'addressed' had higher success rates, emphasizing the importance of initialization protocol. The protocol informs the robot that the human's role is to correct errors to resume the assembly task. The case study found that a clear initialization process is crucial for the robot to understand its role and functions within the system.

There were also limitations in the case study and evaluation of the proposed framework. Specifically, we only evaluated a limited range of commands from the operator, rather than assessing the overall system capabilities. These commands were related to predefined assembly scenarios and did not include other interactions, such as human interruptions and task-irrelevant questions from the human. While \gls{llm}-generated error messages were understandable for knowledgeable operators, there is a risk of low-quality messages for less experienced operators or in varied contexts. The case study also did not study the variability in the operator's knowledge, as the framework assumed operators would provide relevant instructions to the assembly task. Furthermore, the operator was not allowed to change the defined tasks, e.g., variations in positioning or task order.

Future work will analyze the developed framework in various manufacturing scenarios. We will involve users without assembly knowledge, allowing free interaction to complete tasks. Additionally, we will compare the system to other methods by evaluating adaptability, error handling, and performance. The framework will also be assessed using different initialization protocols and tested on various manufacturing assembly tasks.

\section{Conclusion and Future Work}
\label{sec:conclusion}

Advancements in \gls{llm} are applied to human-robot collaborative assembly to execute actions and collaborate based on environmental data. Incorporating \gls{llm} allows robots to better understand human operators, resolve errors, and improve execution with environmental feedback. Based on these insights, we integrated \gls{llm} into our assembly framework for dynamic responses to task variations.

This research addresses key challenges in human-robot collaborative assembly, including developing communication systems that require minimal robotics training \textit{(C1)}, enhancing adaptability to manage changes and errors \textit{(C2)}, and integrating advanced technologies with a human-centric design to improve usability \textit{(C3)}. We validated our framework using the cable shark device assembly process, demonstrating its ability to facilitate intuitive human-robot communication via voice commands with language variations. We dynamically adapt to task variations and errors by integrating \gls{llm}, sensors, and task control mechanisms, demonstrating its ability to maintain productivity and ensure a continuous workflow.

Our next steps will test the framework under real industrial conditions, including operator variations and different manufacturing environments (e.g., noise, dust, brightness). Additionally, future work includes extending the \gls{llm}-based framework to increase adaptability by feeding the \gls{llm} with diverse data on robotic tasks and sensor information. This will enhance the robot's task flexibility, safety, and ability to handle unexpected errors. We also plan to incorporate multiple modalities, such as haptic and gestures to improve human-robot interaction. By integrating \gls{llm} to this multimodal strategy, we aim to improve communication efficiency, task adaptability, and intuitive interaction in the manufacturing environment.

\section*{Acknowledgment}
We thank Dana Smith from DMI Companies, Inc. for providing the case study and offering valuable feedback on the project.

\balance


\end{document}